\pdfoutput=1
\documentclass{isprs} 
\usepackage{subfigure}
\usepackage{setspace}
\usepackage{geometry} 
\usepackage{epstopdf}
\usepackage[labelsep=period]{caption}  
\usepackage[none]{hyphenat}
\usepackage{breakcites} 
\usepackage{flushend}

\begin{document}
\begin{center}
This paper has been published by the ISPRS in the Int. Arch. Photogramm. Remote Sens. Spatial Inf. Sci., volume XLII-2/W10:\break
\\Agrafiotis, P., Skarlatos, D., Georgopoulos, A., and Karantzalos, K.: SHALLOW WATER BATHYMETRY MAPPING FROM UAV IMAGERY BASED ON MACHINE LEARNING, \textit{Int. Arch. Photogramm. Remote Sens. Spatial Inf. Sci.}, XLII-2/W10, 9–16, https://doi.org/10.5194/isprs-archives-XLII-2-W10-9-2019, 2019.\break
\break
\break
The work described here has been extended and published in the following articles:\break
\\Agrafiotis, P.; Karantzalos, K.; Georgopoulos, A.; Skarlatos, D. Correcting Image Refraction: Towards Accurate Aerial Image-Based Bathymetry Mapping in Shallow Waters. \textit{Remote Sens}. 2020, 12, 322. https://doi.org/10.3390/rs12020322\break
\\Agrafiotis, P.; Skarlatos, D.; Georgopoulos, A.; Karantzalos, K. DepthLearn: Learning to Correct the Refraction on Point Clouds Derived from Aerial Imagery for Accurate Dense Shallow Water Bathymetry Based on SVMs-Fusion with LiDAR Point Clouds. \textit{Remote Sens}. 2019, 11, 2225. https://doi.org/10.3390/rs11192225
\end{center}
\pagebreak

\title{SHALLOW WATER BATHYMETRY MAPPING FROM UAV IMAGERY BASED ON MACHINE LEARNING}

\author{
 P. Agrafiotis \textsuperscript{1,2}\thanks{Corresponding author, email: pagraf@central.ntua.gr} , D. Skarlatos \textsuperscript{2}, A. Georgopoulos \textsuperscript{1}, K. Karantzalos \textsuperscript{1}}

\address{
	\textsuperscript{1 }National Technical University of Athens, School of Rural and Surveying Engineering, Department of Topography,
Zografou Campus, \\9 Heroon Polytechniou str., 15780, Athens, Greece
 \\(pagraf, drag, karank)@central.ntua.gr\\

	\textsuperscript{2 }Cyprus University of Technology, Civil Engineering and Geomatics Dept., Lab of Photogrammetric Vision,
\\2-8 Saripolou str., 3036, Limassol, Cyprus
 \\(panagiotis.agrafioti, dimitrios.skarlatos)@cut.ac.cy
}


\commission{II, }{II} 
\workinggroup{II/9} 
\icwg{}   

\abstract{
The determination of accurate bathymetric information is a key element for near offshore activities, hydrological studies such as coastal engineering applications, sedimentary processes, hydrographic surveying as well as archaeological mapping and biological research. UAV imagery processed with Structure from Motion (SfM) and Multi View Stereo (MVS) techniques can provide a low-cost alternative to established shallow seabed mapping techniques offering as well the important visual information. Nevertheless, water refraction poses significant challenges on depth determination. Till now, this problem has been addressed through customized image-based refraction correction algorithms or by modifying the collinearity equation. In this paper, in order to overcome the water refraction errors, we employ machine learning tools that are able to learn the systematic underestimation of the estimated depths. In the proposed approach, based on known depth observations from bathymetric LiDAR surveys, an SVR model was developed able to estimate more accurately the real depths of point clouds derived from SfM-MVS procedures. Experimental results over two test sites along with the performed quantitative validation indicated the high potential of the developed approach.}


\maketitle


\section{INTRODUCTION}\label{INTRODUCTION}
\sloppy
Although through-water depth determination from aerial imagery is a much more time consuming and costly process, it is still a more efficient operation than ship-borne sounding methods and underwater photogrammetric methods \cite{Agrafiotis2018} in the shallower (less than 10 m depth) clear water areas. Additionally, a permanent record is obtained of other features in the coastal region such as tidal levels, coastal dunes, rock platforms, beach erosion, and vegetation. This is true, even though many alternatives for bathymetry \cite{Menna2018} have arose since. This is especially the case for the coastal zone of up to 10m depth, which concentrates most of the financial activities, is prone to accretion or erosion, and is ground for development, where there is no affordable and universal solution for seamless underwater and overwater mapping. Image-based techniques fail due to wave breaking effects and water refraction, and echo sounding fails due to short distances.

At the same time bathymetric LiDAR with simultaneous image acquisition is a valid, albeit expensive alternative, especially for small scale surveys. In addition, despite the fact that the image acquisition for orthophotomosaic generation in land is a solid solution, the same cannot be said for the shallow water seabed. Despite the accurate and precise depth map provided by LiDAR, the sea bed orthoimage generation is prohibited due to the refraction effect, leading to another missed opportunity to benefit from a unified seamless mapping process.
\newpage            
\subsection{Description of the problem}\label{sec:Description of the problem}
Even though UAVs are well established in monitoring and 3D recording of dry landscapes and urban areas, when it comes to bathymetric applications, errors are introduced due to the water refraction.  Unlike in-water photogrammetric procedures where, according to the literature \cite{Lavest2000}, thorough calibration is sufficient to correct the effects of refraction, in through-water (two-media) cases, the sea surface undulations due to waves \cite{Fryer1985,Okamoto1982} and the magnitude of refraction that differ at each point of every image, lead to unstable solutions \cite{Agrafiotis2015,Georgopoulos2012}. More specifically, according to Snell’s law, the effect of refraction of a light beam to water depth is affected by water depth and angle of incidence of the beam in the air/water interface. The problem becomes even more complex when multi view geometry is applied.  
\begin{figure}[ht!]
\begin{center}		\includegraphics[width=0.95\columnwidth]{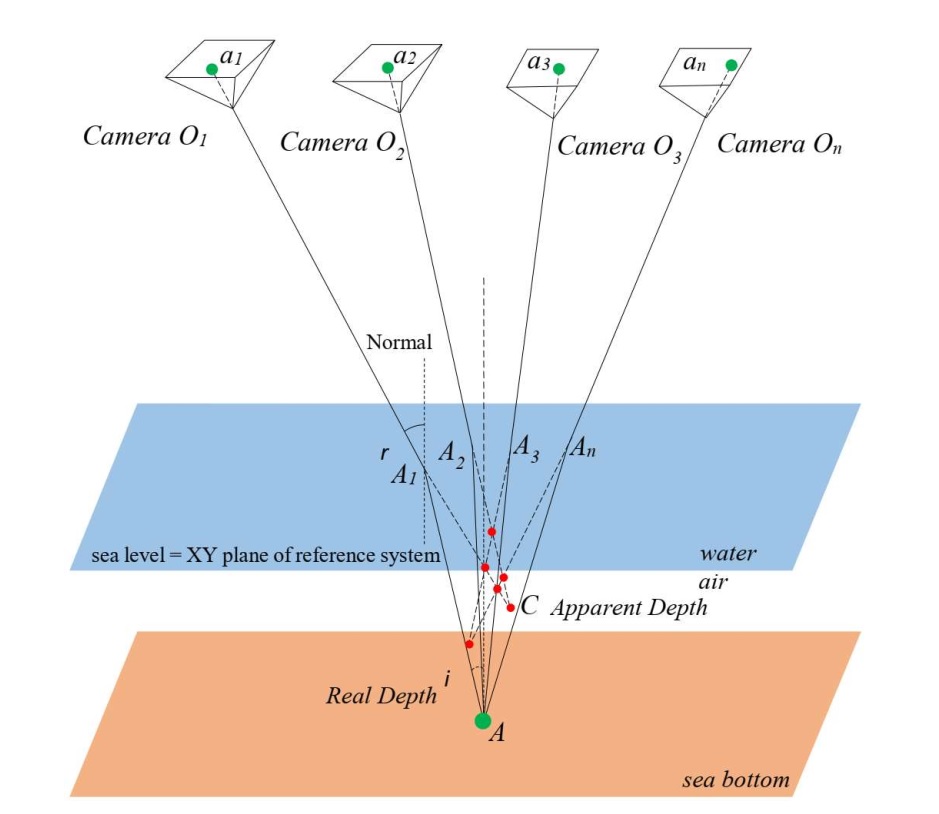}
	\caption{The geometry of two-media photogrammetry for the multiple view case}
\label{fig:figure1}
\end{center}
\end{figure}
In Figure \ref{fig:figure1} the multiple view geometry which applies to the UAV imagery is demonstrated: there, the apparent depth \textit{C} is calculated by the collinearity equation. Starting from the apparent (erroneous) depth of a point \textit{A}, its image-coordinates $a_{1}$, $a_{2}$, $a_{3}$…, $a_{n}$, can be backtracked in images $O_{1}$, $O_{2}$ , $O_{3}$, …, $O_{n}$ using the standard collinearity equation. If a point has been matched successfully in the photos $O_{1}$, $O_{2}$ , $O_{3}$, …, $O_{n}$, then the standard collinearity intersection would have returned the point \textit{C}, which is the apparent and shallower position of point A and in the multiple view case is the adjusted position of all the possible red dots in Figure \ref{fig:figure1}, which are the intersections for each stereopair.
Thus, without some form of correction, refraction produce an image and consequently a point cloud of the submerged surface which appears to lie at a shallower depth than the real surface. In literature, two main approaches to correct refraction in through-water photogrammetry can be found; analytical or image based.

In this work, a new approach to address the systematic refraction errors of point clouds derived from SfM-MVS procedures is introduced. The developed technique is based on machine learning tools which are able to accurately recover shallow bathymetric information from UAV-based imaging datasets, leveraging several coastal engineering applications. In particular, the goal was to deliver image-based point clouds with accurate depth information by learning to estimate the correct depth from the systematic differences between image-based products and (the current gold-standard for shallow waters) LiDAR point clouds. To this end, a Linear Support Vector Regression model was employed and trained to predict the actual depth \textit{Z} from the apparent depth of a point, $Z_{o}$ from the image-based point cloud.
The rest of the paper is organized as follows: Subsection \ref{sec:Related work} presents the related work regarding refraction correction and the use of SVMs in bathymetry determination. In Section \ref{sec:DATASETS}, datasets used are described while in Section \ref{sec:PROPOSED METHODOLOGY} the proposed methodology is described and justified. In Section \ref{sec:TESTS AND EVALUATION} the tests performed and the evaluations carried out are described. Section \ref{sec:CONCLUSIONS} concludes the paper.
\subsection{Related work}\label{sec:Related work}
Refraction effect has driven scholars to suggest several models for two-media photogrammetry, most of which are dedicated to specific applications. Two-media photogrammetry is divided into through-water and in-water photogrammetry. The through-water term is used when the camera is above the water surface and the object is underwater, hence part of the ray is traveling through air and part of it through water. It is most commonly used in aerial photogrammetry \cite{Skarlatos2018,Dietrich2017} or in close range applications \cite{Georgopoulos2012,Butler2002}. It is argued that if the water depth to flight height ratio is considerably low, then water refraction is unnecessary. However, as shown in the literature \cite{Skarlatos2018}, the water depth to flying height ratio is irrelevant, in cases ranging from drone and unmanned aerial vehicle (UAV) mapping to full-scale manned aerial mapping. In these cases water refraction correction is necessary.
\subsection{Bathymetry Determination using Machine Learning}\label{sec:Bathymetry Determination using Machine Learning}
Even though the presented approach here is the only one dealing with UAV imagery and dense point clouds resulting from the SfM-MVS processing, there is a small number of single image approaches for bathymetry retrieval using satellite imagery.  Most of these methods are based on the relation between the reflectance and the depth. These approaches exploit a support vector machine (SVM) system to predict the correct depth \cite{Wang2018,Misra2018}. Experiments there showed that the localized model reduced the bathymetry estimation error by 60\% from an RMSE of 1.23m to 0.48m.  In \cite{Mohamed2016} a methodology is introduced using an Ensemble Learning (EL) fitting algorithm of Least Squares Boosting (LSB) for bathymetric maps calculation in shallow lakes from high resolution satellite images and water depth measurement samples using Echo-sounder. The retrieved bathymetric information from the three methods was evaluated using Echo Sounder data. The LSB fitting ensemble resulted in an RMSE of 0.15m where the PCA and GLM yielded RMSE’s of 0.19m and 0.18m respectively over shallow water depths less than 2m. Except from the primary data used, the main difference between the work presented here and the work presented in these articles, is that they test and evaluate their proposed algorithms on percentages of the same test site and at very shallow depths while here two different test sites are used.
\section{DATASETS}\label{sec:DATASETS}
The proposed methodology has been applied in real-world applications in two different test sites for verification and comparison against bathymetric LiDAR data. In the following paragraphs, the results of the proposed methodology are investigated and evaluated. The initial point cloud used here can be created by any commercial photogrammetric software (such as Agisoft’s Photoscan©, used in this study) following standard process, without water refraction compensation. However, wind affects the sea surface with wrinkles and waves. Taking this into account, the water surface needs to be as flat as possible, so that to have best sea bottom visibility and follow the assumption of flat-water surface. In case of a wavy sea surface, errors would be introduced \cite{Okamoto1982,Agrafiotis2015} without any form of correction \cite{Chirayath2016} applied and the relation of the real and the apparent depths will be more scattered, affecting to some extent the training and the fitting of the model. Furthermore, water should not be turbid enough to have a clear bottom view. Obviously, water turbidity and water visibility are additional restraining factors. Just like in any photogrammetric project, sea bottom must present pattern, meaning that photogrammetric bathymetry might fail in sandy or seagrass sea bed. However, since normally, a sandy bottom does not present any abrupt height differences and detailed forms, and provided measures to eliminate the noise of the point cloud in these areas are taken, results would be acceptable, even in a less dense point cloud, due to matching difficulties.  
\subsection{Test sites and available data}\label{sec:Test sites and available data}
In order to facilitate the training and the testing of the proposed approach, ground truth data of the seabed depth were required, together with the image-based point clouds. To facilitate this, ground control points (GCPs) were measured in land and used to georeference the photogrammetric data with the LiDAR data. The common system used is the Cyprus Geodetic Reference System (CGRS) 1993, to which the LiDAR data were already georeferenced.
\subsubsection{Amathouda Test Site}\label{sec:Amathouda Test Site}
The first site used is Amathouda (Figure \ref{fig:figure2} upper image), where the seabed reaches a maximum depth of 5.57 m. The flight was executed with a Swinglet CAM fixed-wing UAV with an Canon IXUS 220HS camera having 4.3mm focal length, 1.55$\mu$m pixel size and 4000$\times$3000 pixels format. A total of 182 photos were acquired, from an average flight height of 103 m, resulting in 3.3 cm average GSD.  
\subsubsection{Agia Napa Test Site}\label{sec:Agia Napa Test Site}
\begin{figure}[ht!]
\begin{center}
		\includegraphics[width=0.95\columnwidth]{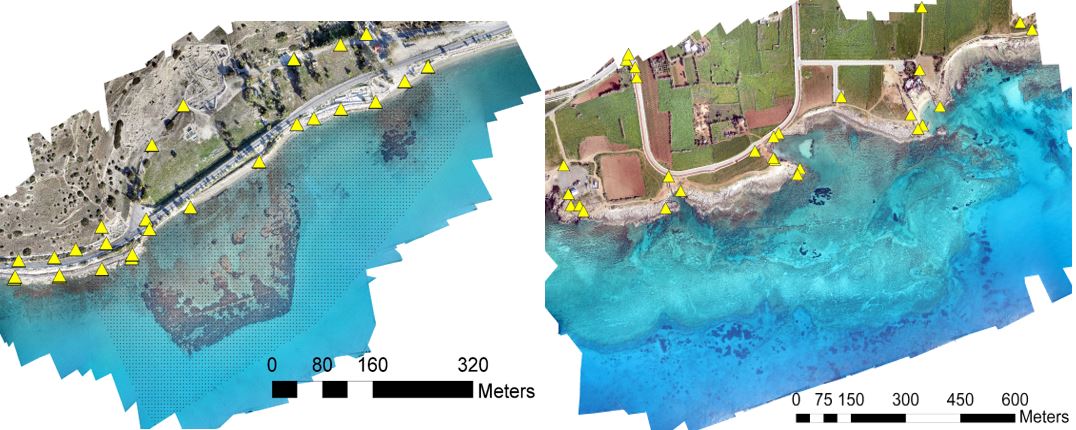}
	\caption{The two test sites. Amathouda (top) and Ag. Napa (bottom). Yellow triangles represent the GCPs positions.}
\label{fig:figure2}
\end{center}
\end{figure}

The second test site is in Agia Napa (Figure \ref{fig:figure2} lower image), where the seabed reaches the depth of 14.8m. The flight here executed with the same UAV. In total 383 images were acquired, from an average flight height of 209m, resulting in 6.3cm average ground pixel size. 
Table \ref{fig:t1}(presents the flight and image-based processing details of the two different test sites. There, it can be noticed that the two sites have a different average flight height, indicating that the suggested solution is not limited to specific flight heights. That means that a trained model on an area may be applied on another area, having the flight and image-based processing characteristics of the datasets used.
\begin{table}[ht!]
\begin{center}		\includegraphics[width=1.0\columnwidth]{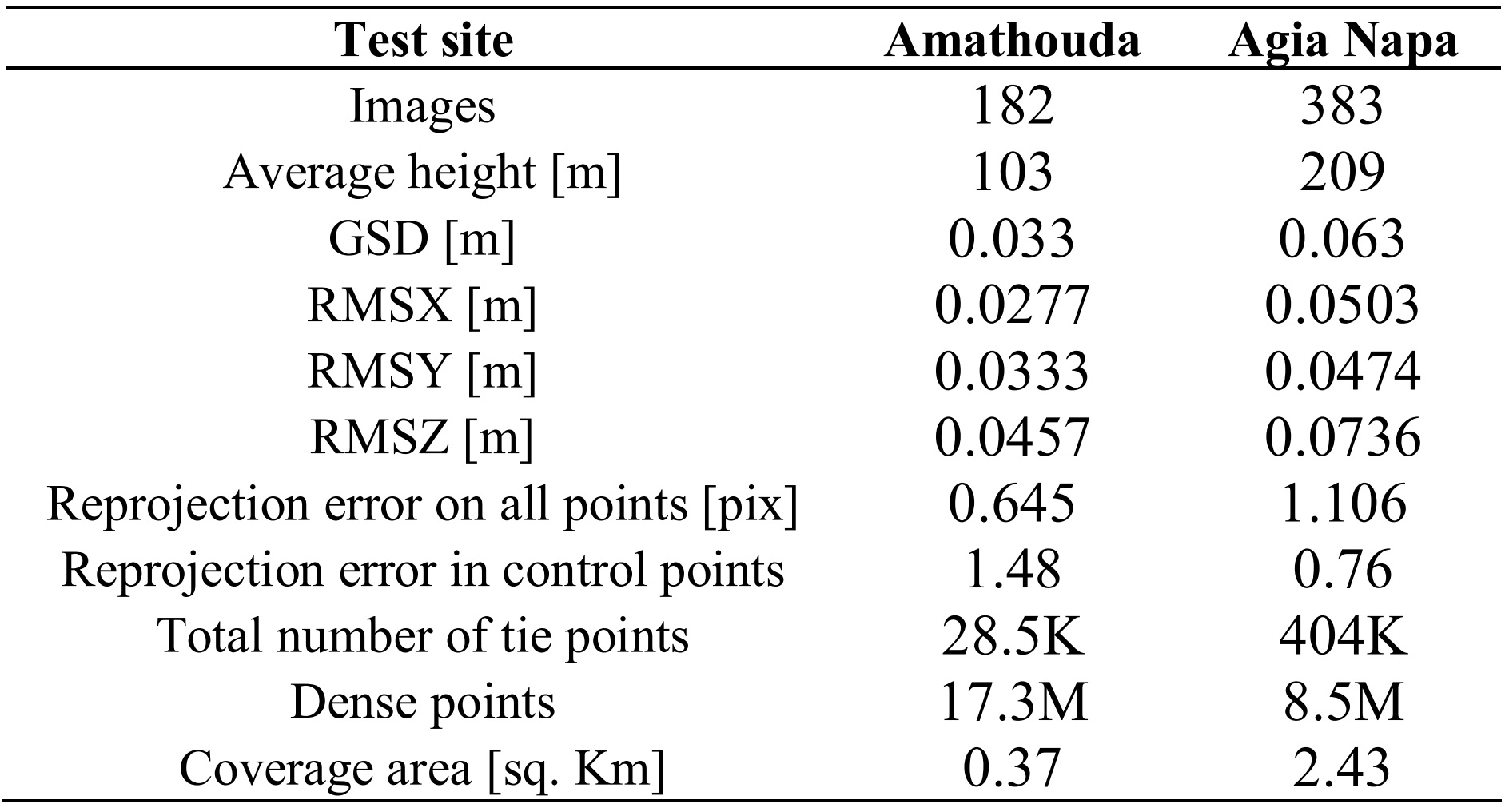}
	\caption{Flight and image-based processing details regarding the two different test sites}
\label{fig:t1}
\end{center}
\end{table}
\subsubsection{Data pre-processing}\label{sec:Data pre-processing}
\begin{figure*}[ht!]
\begin{center}
		\includegraphics[width=1.85\columnwidth]{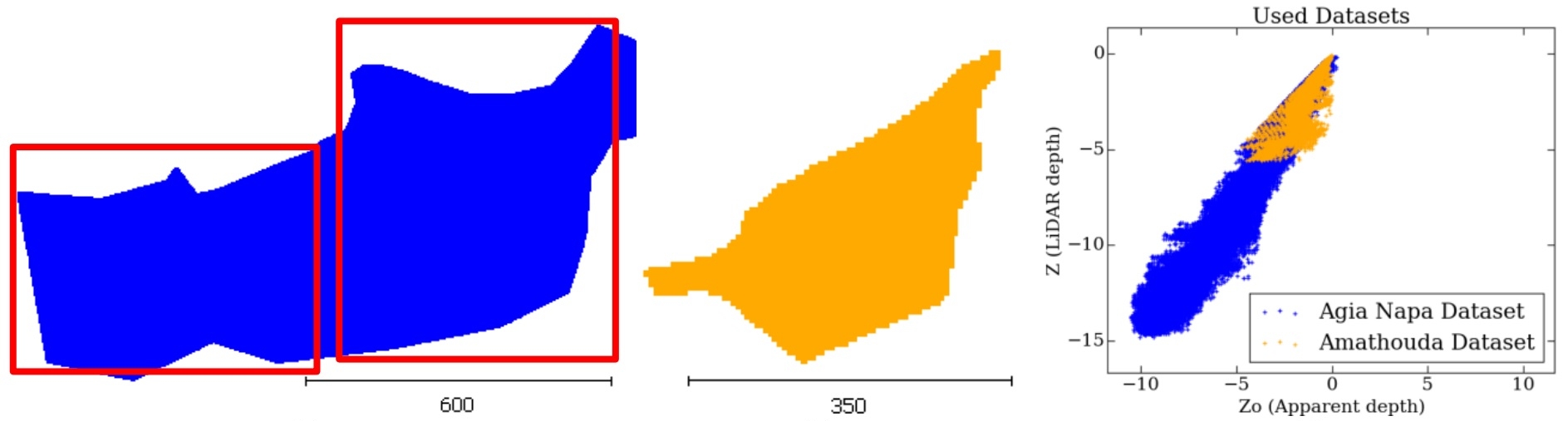}
	\caption{The two test areas from the Agia Napa test site are presented (left) with blue colour: Part I on the left and Part II on the right. The Amathouda test site is presented in the middle with orange colour. The distribution of the \textit{Z} and $Z_{o}$ values for each dataset is presented (right) as well.}
\label{fig:figure3}
\end{center}
\end{figure*}

To facilitate the training of the proposed bathymetry correction model, data were pre-processed. Since the image-based point cloud was denser, than the LiDAR point cloud, it was decided to reduce the number of the points of the first one. To that direction the number of the image-based point clouds were reduced to the number of the LiDAR point clouds, for the two test sites. This way, for each position \textit{X}, \textit{Y} of the seabed two depths are corresponding: the apparent depth $Z_{o}$ and the LiDAR depth \textit{Z}. Consequently, outlier data were removed from the dataset. At this stage of the pre-processing, outliers were considered points having $Z_{o}$ $\geq$ \textit{Z} since this is not valid when the refraction phenomenon is present. Moreover, points having $Z_{o}$ $\geq$ 0m were also removed since they might cause errors in the training process.  After being pre-processed, the datasets were used as follows: due to availability of a lot of reference data in Agia Napa test site, the site was split in two parts having different characteristics: Part I having 627.522 points (Figure \ref{fig:figure3}(left) in the red rectangle on the left, Figure \ref{fig:figure5}(top left)) and Part II having 661.208 points (Figure \ref{fig:figure3}(left) in the red rectangle on the right, Figure \ref{fig:figure5}(top right)). Amathouda dataset (Figure \ref{fig:figure3}(middle) and Figure \ref{fig:figure5}(bottom left)) was not split since the available points were much less and quite scattered (Figure \ref{fig:figure3}(right)). The distribution of the \textit{Z} and $Z_{o}$ of the points is presented in Figure \ref{fig:figure3}(right) the Agia Napa dataset is presented with blue colour, while the Amathouda dataset is presented with orange colour.
\subsubsection{LiDAR Reference data}\label{sec:LiDAR Reference data}
LiDAR point clouds of the submerged areas were used as reference data for training and evaluation of the developed methodology. These point clouds were generated with the Leica HawkEye III (Leica Geosystems AG, Heerbrugg, Switzerland), a deep penetrating bathymetric airborne LiDAR system. This multisensory system includes a 10 kHz bathymetric channel (532 nm) for deep water and a 35 kHz bathymetric channel, optimized for shallow water and the transition zone towards the shore. Table \ref{fig:t3} presents the details of the LiDAR data used.
\begin{table}[ht!]
\begin{center}
		\includegraphics[width=1.0\columnwidth]{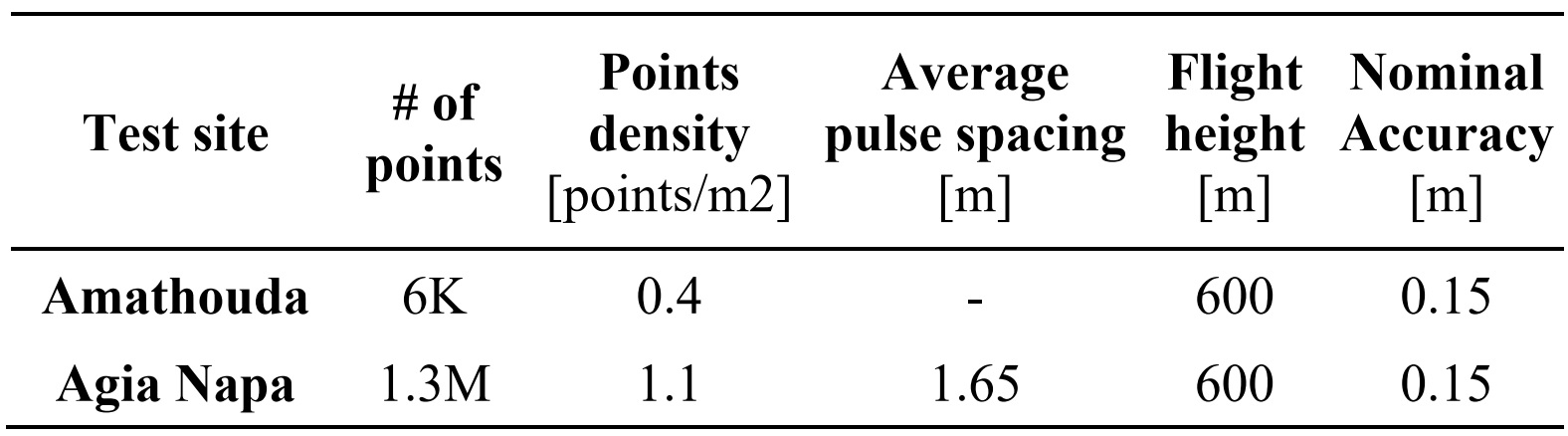}
	\caption{LiDAR data specifications}
\label{fig:t2}
\end{center}
\end{table}
Even though the specific LiDAR system can offer point clouds with accuracy of 20mm in topographic applications according to the manufacturers, when it comes to bathymetric applications the system’s error range is in the order of +/-150mm for depths up to 1.5 Secchi depth, similar to other conventional topographic airborne scanners \cite{Steinbacher2012}. According to the literature LiDAR bathymetry data can be affected by significant systematic errors that lead to much greater errors. In \cite{Skinner2011} the average error in elevations for the wetted river channel surface area was -0.5\% and ranged from -12\% to 13\%. In \cite{Bailly2010} authors detected a random error of 0.19m-0.32m for the riverbed elevation from the Hawkeye II sensor. In \cite{Fernandez2014} the standard deviation of the bathymetry elevation differences calculated reaches 0.79m, with 50\% of the differences falling between 0.33m to 0.56m. However, according to the authors it appears that most of these differences are due to sediment transport between observation epochs. In \cite{Westfeld2017} authors report that the RMSE of the lateral coordinate displacement is 2.5\% of the water depth for the smooth, rippled sea swell. Assuming a mean water depth of 5m leads to a RMSE of 12cm. If a light sea state with small wavelets assumed, results with an RMSE of 3.8\% which corresponds to 19cm in 5m water are expected. It becomes obvious that wave patterns can cause significant systematic effects in bottom coordinate locations. Even for very calm sea states, the lateral displacement can be up to 30cm at 5m water depth \cite{Westfeld2017}. 

Considering the above, authors would like to highlight here that in the proposed approach, LiDAR point clouds are used for training the suggested model, since this is the State-of-the-Art method used for shallow water bathymetry of large areas \cite{Menna2018}, even though in some cases the absolute accuracy of the resulting point clouds is deteriorated. These issues do not affect the principle of the main goal of the presented approach which is to systematically solve the depth underestimation problem, by predicting the correct depth, as proved in the next sections.
\section{PROPOSED METHODOLOGY}\label{sec:PROPOSED METHODOLOGY}
A Support Vector Regression (SVR) method is adopted in order to address the described problem. To that direction, data available from two different test sites, characterized by different type of seabed and depths are used to train, validate and test the proposed approach. The Linear SVR model was selected after studying the relation of the real (\textit{Z}) and the apparent ($Z_{o}$) depths of the available points (Figure \ref{fig:figure3}(right)). Based on the above, the SVR model fits according to the given training data: the LiDAR (\textit{Z}) and the apparent depths ($Z_{o}$) of many 3D points. After fitting, the real depth can be predicted in the cases where only the apparent depth is available. In the performed study the relationship of the LiDAR (\textit{Z}) and the apparent depths ($Z_{o}$) of the available points rather follows a linear model and as such, a deeper learning architecture was not considered necessary.
\begin{figure}[ht!]
\begin{center}
		\includegraphics[width=1.0\columnwidth]{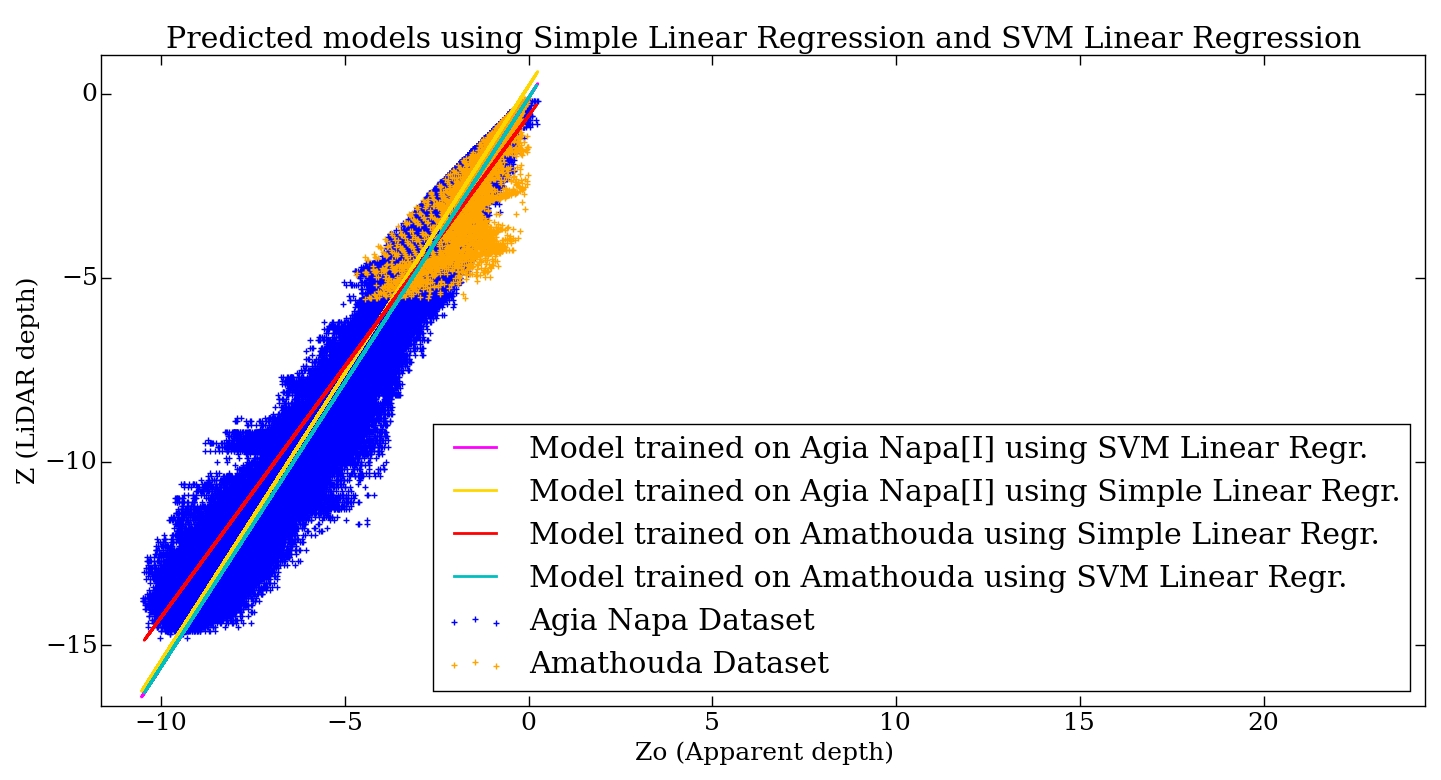}
	\caption{The established correlations based on a simple Linear Regression and SVM Linear Regression models, trained on Amathouda and Agia Napa datasets.}
\label{fig:figure4}
\end{center}
\end{figure}
The use of a simple Linear Regression model was also examined, fitting tests were performed in the two test sites and predicted values were compared to the LiDAR data. However, this approach was rejected since the predicted models were producing larger errors than the ones produced by the SVM Linear Regression and they were highly dependent on the training dataset and its density, being very sensitive to the noise of the point cloud. This is explained by the fact that the two regression methods differ only in the loss function where SVM minimizes hinge loss while logistic regression minimizes logistic loss and logistic loss diverges faster than hinge loss being more sensitive to outliers. This is apparent also in Figure \ref{fig:figure4}, where the predicted models using a simple Linear Regression and an SVM Linear Regression trained on Amathouda and Agia Napa [I] datasets are plotted. In the case of training on the Amathouda dataset, it is obvious that the two predicted models (lines in red and cyan colour) differ considerably as the depth increases, leading to different depth predictions. However, in the case of the models trained in Agia Napa [I] dataset, the two predicted models (lines in magenta and yellow colour) are overlapping, also with the predicted model of the SVM Linear Regression, trained on Amathouda. These results suggest that the SVM Linear Regression is less dependent on the density and the noise of the data and ultimately the more robust method, predicting systematically reliable models, outperforming simple Linear Regression.
\subsection{Linear SVR}\label{sec:Linear SVR}
SVMs can also be applied to regression problems by the introduction of an alternative loss function \cite{Smola1996}. The loss function must be modified to include a distance measure. In this paper, a Linear Support Vector Regression model is used exploiting the implementation of \cite{Pedregosa2011}. The problem is formulated as follows: consider the problem of approximating the set of depths:
\begin{equation}
D = \{(Z_{0}^1, Z^1), ..., (Z_{0}^l, Z^l)\}, \hspace{5mm} Z_{0} \in R^n, \hspace{5mm} Z \in R
\label{equ:1}
\end{equation} 
with a linear function
\begin{equation} 
f(Z_{0}) = \langle w,Z_{0} \rangle + b
\label{equ:2}
\end{equation} 
The optimal regression function is given by the minimum of the functional,
\begin{equation}
\phi(w,Z_{0}) = \frac{1}{2}\|w\|^2+c\sum_{i}(\xi_{i}^- + \xi_{i}^+)
\label{equ:3}
\end{equation} 
Where \textit{c} is a pre-specified positive numeric value that controls the penalty imposed on observations that lie outside the epsilon margin ($\varepsilon$) and helps to prevent overfitting (regularization). This value determines the trade-off between the flatness of \textit{f}($Z_{o}$) and the amount up to which deviations larger than $\varepsilon$ are tolerated, and $\xi_i-$, $\xi_i+$ are slack variables representing upper and lower constraints of the outputs of the system, \textit{Z} is the real depth of a point \textit{X}, \textit{Y} and $Z_{o}$ is the apparent depth of the same point \textit{X}, \textit{Y}. Based on the above, the proposed framework is trained using the real (\textit{Z}) and the apparent ($Z_{o}$) depths of a number of points in order to predict the real depth in the cases where only the apparent depth is available.
\section{TESTS AND EVALUATION}\label{sec:TESTS AND EVALUATION}
\subsection{Training, Validation and Testing}\label{sec:Training, Validation and Testing}
In order to evaluate the performance of the developed model in terms of robustness and effectiveness, six different training sets were formed from two test sites of different seabed characteristics and then validated against 13 different testing sets.  
\subsubsection{Agia Napa and Amathouda datasets}\label{sec:Agia Napa and Amathouda datasets}
\begin{figure}
\begin{center}
\includegraphics[width=.23\textwidth]{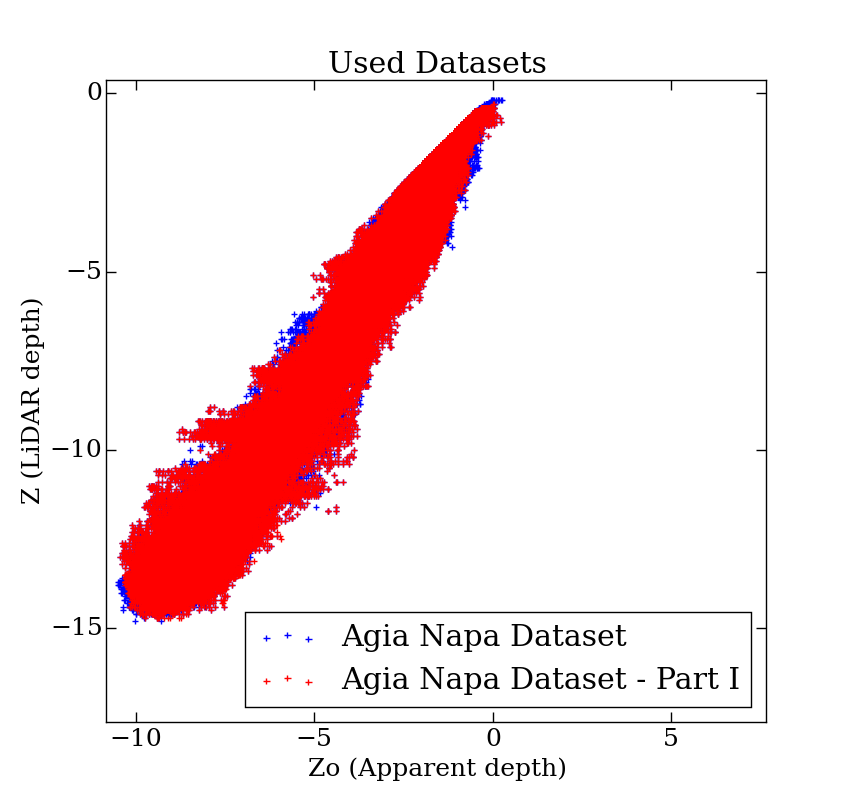}
\includegraphics[width=.23\textwidth]{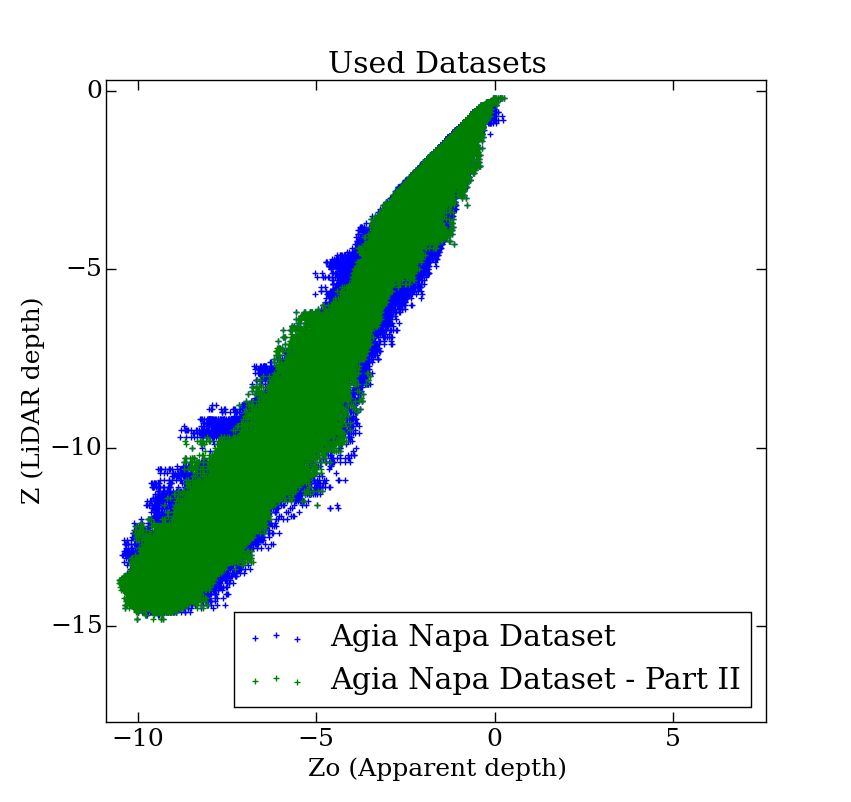}
\includegraphics[width=.23\textwidth]{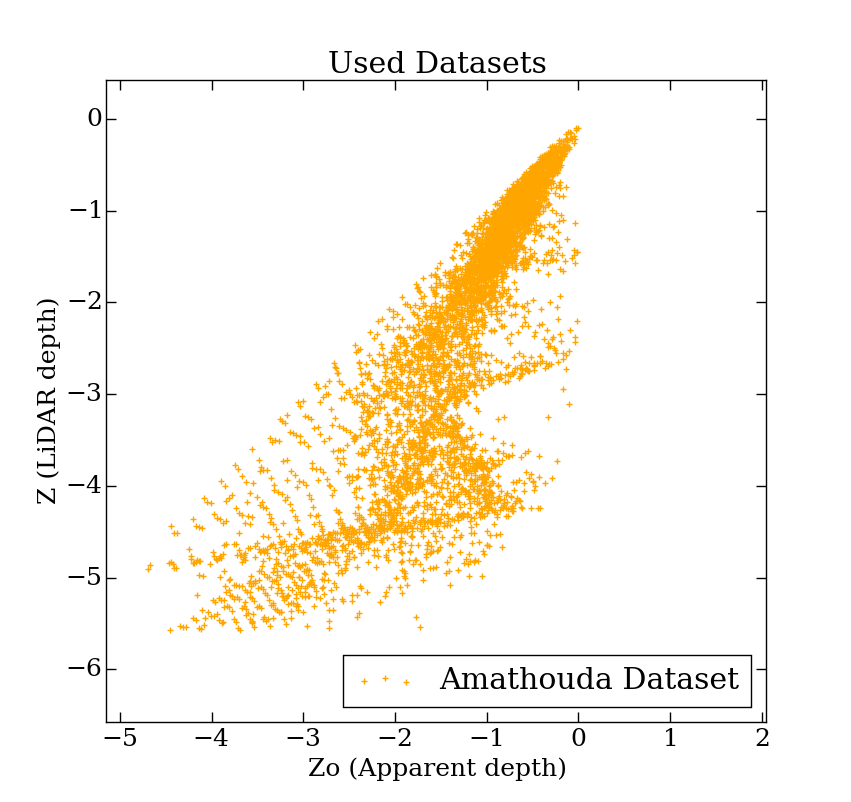}
\includegraphics[width=.23\textwidth]{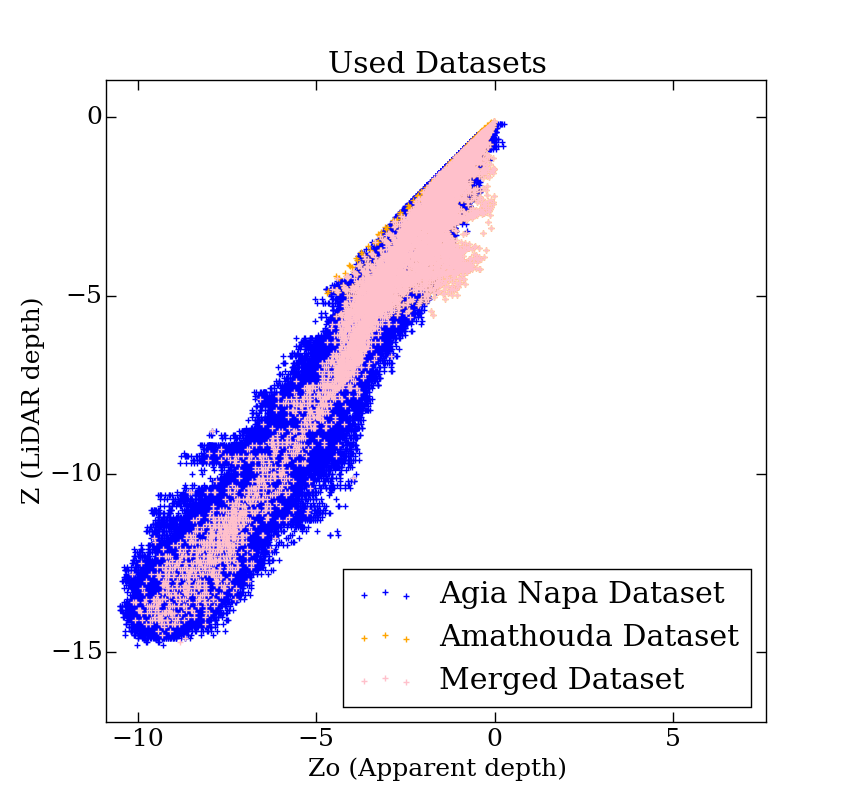}
\caption{The \textit{Z}-$Z_{o}$ distribution of the used datasets:  the Agia Napa Part I dataset over the full Agia Napa dataset (top left), The Agia Napa Part II dataset over the full Agia Napa dataset (top right), Amathouda dataset (bottom left), The merged dataset over the Agia Napa and Amathouda datasets (bottom right).}
\label{fig:figure5}
\end{center}
\end{figure}

The first and the second training approaches are using 5\% and 30\% of the Agia Napa Part II dataset respectively in order to fit the Linear SVR model and predict the correct depth over the Agia Napa Part I and Amathouda test sites. The third and the fourth training approaches are using 5\% and 30\% of the Agia Napa Part I dataset respectively in order to fit the Linear SVR model and predict the correct depth over the Agia Napa Part II and Amathouda test sites. The fifth training approach is using 100\% of the Amathouda dataset in order to fit the Linear SVR model and predict the correct depth over the Agia Napa Part I, the Agia Napa Part II and their combination. The \textit{Z}-$Z_{o}$ distribution of the points used for this training can be seen in Figure \ref{fig:figure5}(bottom left). It is important to notice here that the maximum depth of the training dataset is 5.57m while the maximum depth of the testing datasets is 14.8m and 14.7m respectively.
\subsubsection{Merged dataset}\label{Merged dataset}
Finally, a sixth training approach is performed by creating a virtual dataset containing almost the same number of points from each of these two datasets. The \textit{Z}-$Z_{o}$ distribution of this “merged dataset” is presented in Figure \ref{fig:figure5}(bottom right). In the same figure the \textit{Z}-$Z_{o}$ distribution of the Agia Napa dataset and Amathouda dataset are presented in blue and yellow colour respectively. This dataset was generated using the total of the Amathouda dataset points and 1\% of the Agia Napa Part II dataset.
\subsection{Evaluation of the results}\label{sec:Evaluation of the results}
\begin{figure}[ht!]
\begin{center}
		\includegraphics[width=1.0\columnwidth]{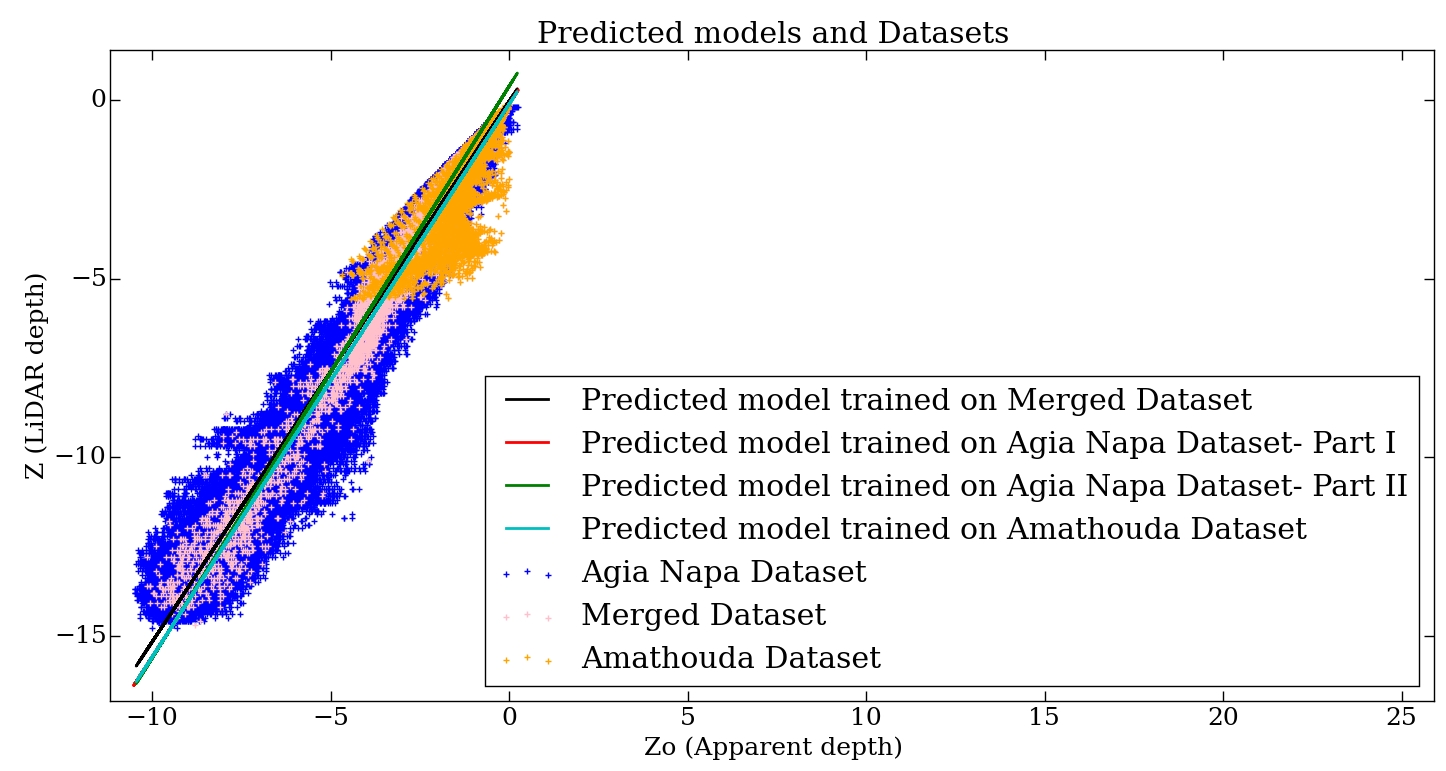}
	\caption{The \textit{Z}-$Z_{o}$ distribution of the employed datasets and the respective predicted linear models}
\label{fig:figure6}
\end{center}
\end{figure}

Figure \ref{fig:figure6} demonstrates four of the predicted models: the black coloured line represents the predicted model trained on the Merged Dataset, the cyan coloured line represents the predicted model trained on the Amathouda Dataset, the red coloured line represents the predicted model trained on the Agia Napa Part I [30\%] Dataset, and the green coloured line represents the predicted model trained on the Agia Napa Part II [30\%] Dataset.
It is obvious that despite the scattered points which lie away from these lines, the models achieve to follow the \textit{Z}-$Z_{o}$ distribution of most of the points. It is important to highlight here that the differences between the predicted model trained on the Amathouda dataset (cyan line) and the predicted models trained on Agia Napa datasets are not remarkable, even though the maximum depth of Amathouda dataset is 5.57m and the maximum depth of Agia Napa datasets is 14.8m and 14.7m respectively. The biggest difference observed between the predicted models is between the predicted model trained on Agia Napa [II] dataset (green line) and the predicted model trained on the Merge dataset (black line): 0.45m at 16.8m depth, or a 2.7\% of the real depth. In the next paragraphs the results of the proposed method are evaluated in terms of cloud to cloud distances. Additionally, cross sections of the seabed are presented to highlight the high performance of the proposed methodology and the issues and differences observed between the tested and ground truth point clouds.
\subsubsection{Multiscale Model to Model Cloud Comparison}\label{sec:Multiscale Model to Model Cloud Comparison}
\begin{figure}[ht!]
\begin{center}
		\includegraphics[width=0.95\columnwidth]{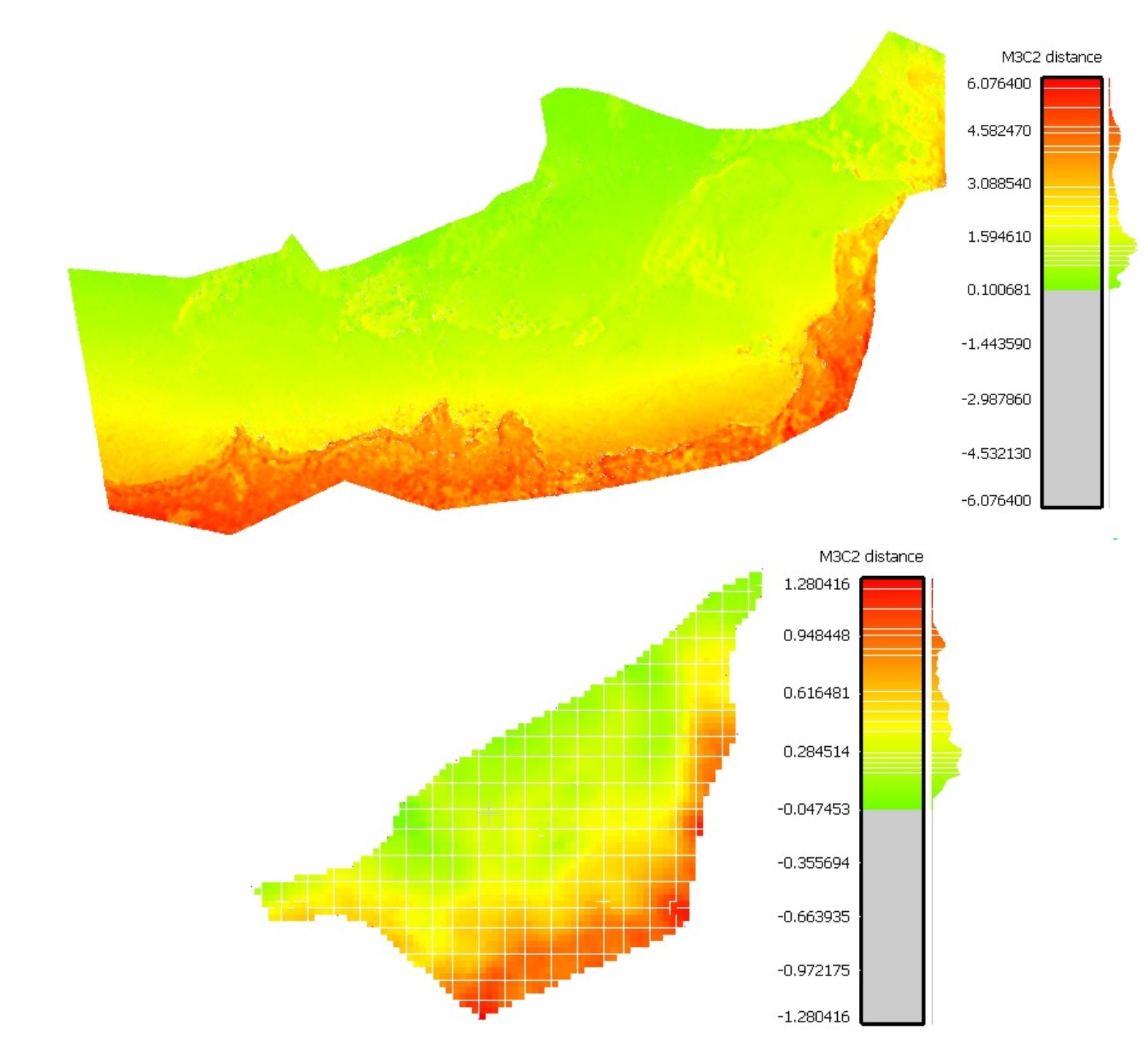}
	\caption{The initial M3C2 distances between the (reference) LiDAR point cloud and the image-based point clouds derived from the SfM-MVS. Figure \ref{fig:figure7}(top) presents the M3C2 distances of Agia Napa and Figure \ref{fig:figure7}(bottom) the initial distances for Amathouda test site.}
\label{fig:figure7}
\end{center}
\end{figure}

\begin{figure*}[ht!]
\begin{center}
		\includegraphics[width=1.9\columnwidth]{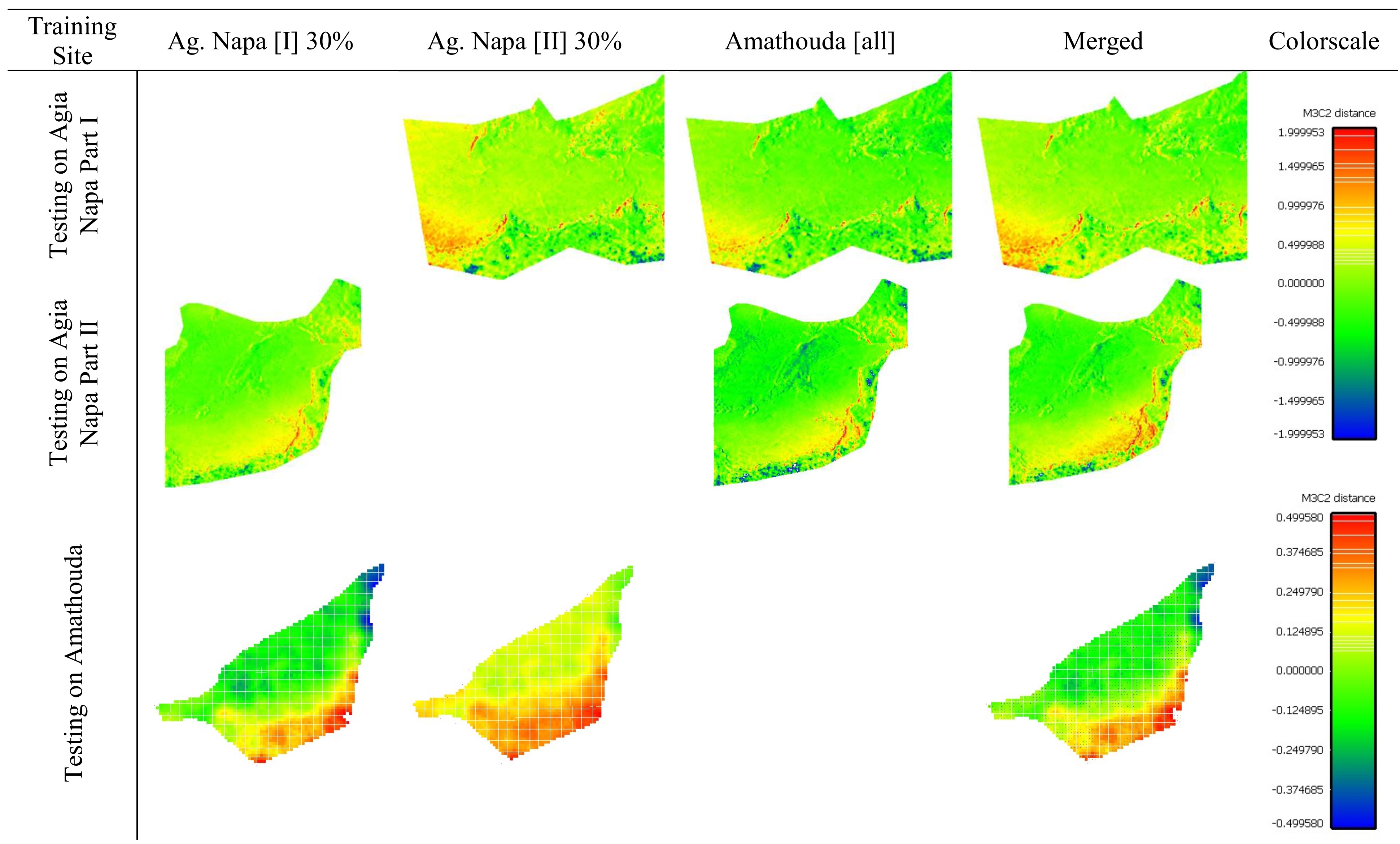}
	\caption{The cloud to cloud (M3C2) distances between the LiDAR point cloud and the recovered point clouds after the application of the proposed approach. The first, the second and the third row of the figure demonstrate the calculated distance maps and their colour scales for the Agia Napa (Part I and Part II) and Amathouda test sites respectively}
\label{fig:figure8}
\end{center}
\end{figure*}

To evaluate the results of the proposed methodology, the initial point clouds of the SfM-MVS procedure and the point clouds resulted from the proposed methodology were compared with the LiDAR point cloud using the Multiscale Model to Model Cloud
Comparison (M3C2) \cite{Lague2013} in Cloud Compare freeware (Cloud Compare, 2019) to demonstrate the changes and the differences that are applied by the presented depth correction approach. The M3C2 algorithm offers accurate surface change measurement that is independent of point density \cite{Lague2013}. In Figure \ref{fig:figure7}(top) and Figure \ref{fig:figure7}(bottom), the distances between the reference data and the original image-based point clouds are increasing as the depth increases. These comparisons make clear that the refraction effect cannot be ignored in such applications. In both cases demonstrated in Figure \ref{fig:figure7}(top) and Figure \ref{fig:figure7}(bottom), the Gaussian mean of the differences is significant reaching 0.44 m (RMSE 0.51m) in the Amathouda test site and 2.23m (RMSE 2.64m) in the Agia Napa test site. Since these values might be considered ‘negligible’ in some applications, it is important to stress that in the Amathouda test site more than 30\% of the compared image-based points present a difference of 0.60-1.00m from the LiDAR points, while in Agia Napa, the same percentage presents differences of 3.00-6.07m, i.e. 20\% - 41.1\% percent of the real depth.
Figure \ref{fig:figure8} presents the cloud to cloud distances (M3C2) between the LiDAR point cloud and the point clouds resulted from the predicted model trained on each dataset. Table \ref{fig:t3} presents the results of each one of the 13 tests performed with every detail. There, a great improvement is observed. More specifically, in Agia Napa [I] test site, the initial 2.23m mean distance is reduced to -0.10m while in Amathouda, the initial mean distance of 0.44m is reduced to -0.03m, including outlier points such as seagrass that are not captured in the LiDAR point clouds for both cases or are caused due to point cloud noise again in areas with seagrass or poor texture. It is important also to note that the large distances between the clouds observed in Figure \ref{fig:figure7} disappear. This improvement is observed in every test performed proving that the proposed methodology based on Machine Learning achieves great reduction of the errors caused by the refraction in the seabed point clouds. In Figure \ref{fig:figure8}, it is obvious that the larger differences between the predicted and the LiDAR depths are observed in some specific areas, or areas with same characteristics. In more detail, the lower-left area of Agia Napa Part I test site and the lower-right area of Agia Napa Part II test site, have constantly larger error than other areas of the same depth. This can be explained by their position in the photogrammetric block, since these are areas far for from the control points, situated in the shore and they are in the outer area of the block. However, it is noticeable that these two areas, present smaller deviation from the LiDAR point cloud, when the model is trained in Amathouda test site, a totally different and shallower test site. Additionally, areas with small rock formations are also presenting large differences. This is attributed to the different level of detail in these areas between the LiDAR point cloud and the image-based one, since LiDAR average point spacing is about 1.1m. These small rock formations in many cases lead M3C2 to detect larger distances in these parts of the site and are responsible for the increased Standard Deviation of the M3C2 distances (Table \ref{fig:t3}).  
\begin{table*}[ht!]
\begin{center}
		\includegraphics[width=1.9\columnwidth]{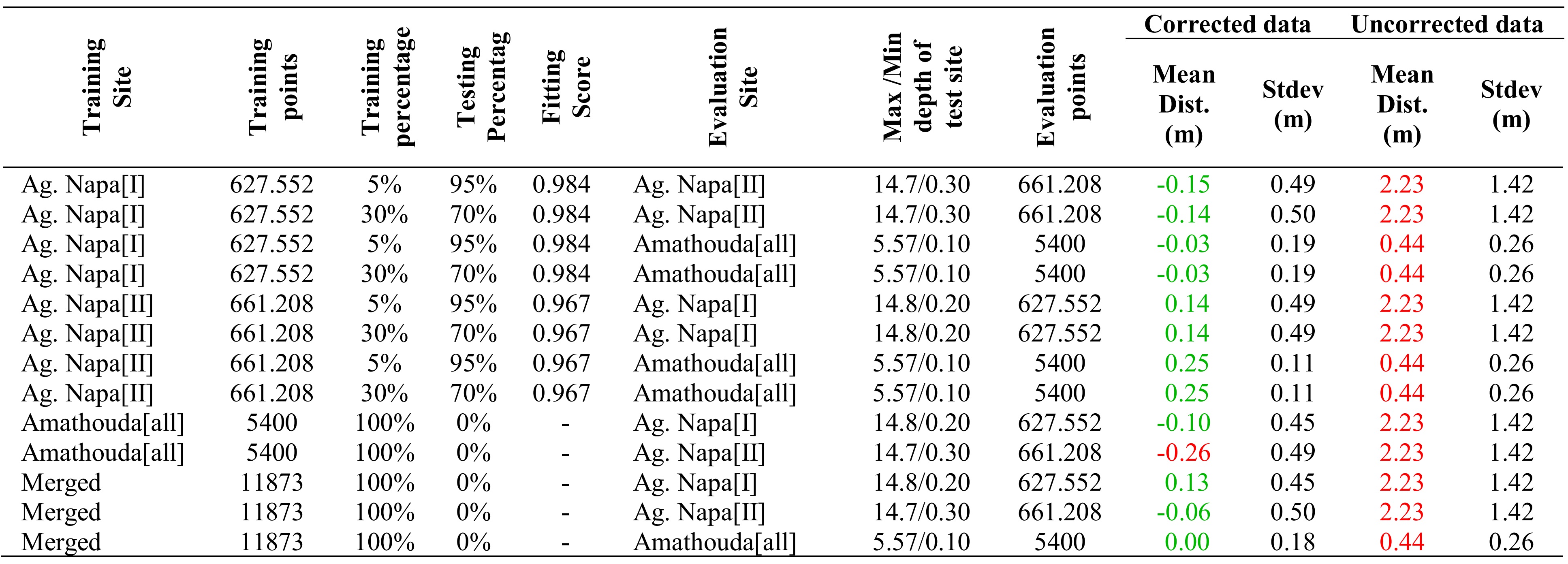}
	\caption{The results of the comparisons between the predicted models for all the tests performed. }
\label{fig:t3}
\end{center}
\end{table*}
\begin{figure*}[ht!]
\begin{center}
\includegraphics[width=1.9\columnwidth]{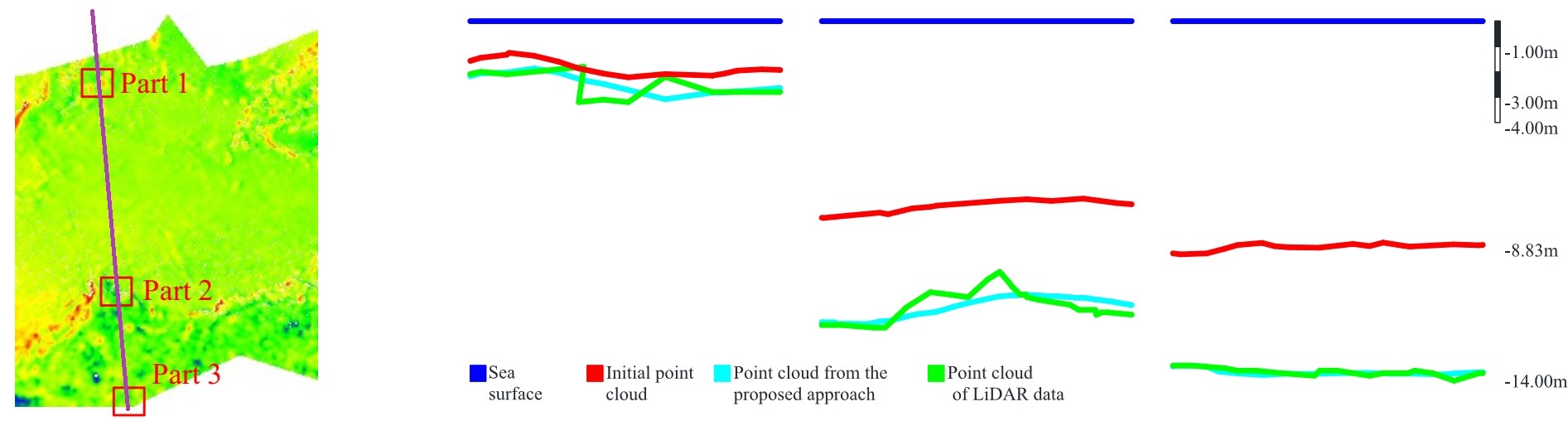}
	\caption{Indicative cross-sections (X and Y axis having the same scale) from the Agia Napa (Part I) region after the application of the proposed approach when trained with 30\% from the Part II region. The blue line corresponds to water surface while the green one corresponds to LiDAR data. The cyan line is the recovered depth after the application of the proposed approach, while the red line corresponds to the depths derived from the initial uncorrected image-based point cloud.}
\label{fig:figure9}
\end{center}
\end{figure*}
\subsubsection{Seabed cross sections}\label{sec:Seabed cross sections}
Several differences observed between the image-based point clouds and the LiDAR data that are not due to the proposed depth correction approach. Cross sections of the seabed were generated with main aim to prove the performance of the proposed method, excluding differences between the compared point clouds. In Figure \ref{fig:figure9} the footprint of a representative cross section is demonstrated together with three parts of the section. These parts highlight the high performance of the algorithm and the differences between the point clouds, reported above. In more detail, in the first and the second part of the section presented, it can be noticed that even if the corrected image-based point cloud is almost matching the LiDAR one on the left and the right side of the sections, in the middle parts, errors are introduced. These are mainly caused by coarse errors which though are not related to the depth correction approach. However, in the third part of the section, it is obvious that even when the depth reaches 14m, the corrected image-based point cloud matches the LiDAR one, indicating a very high performance of the proposed approach. Excluding these differences, the corrected image-based point cloud presents deviations of less than 0.05m (0.36\% remaining error at 14m depth) from the LiDAR point cloud.
\subsubsection{Fitting Score}\label{sec:Seabed cross sections}
Another measure to evaluate the predicted model in cases where a percentage of the dataset has been used for training and the rest percentage has been used for testing is by computing the coefficient $R^{2}$ which is the fitting score and is defined as
\begin{equation}
R^2 = 1- \frac{\sum(Z_{true}-Z_{predicted})^2}{\sum(Z_{true}-Z_{true.mean})^2}
\label{equ:4}
\end{equation}
The best possible score is 1.0 and it can also be negative (Pedregosa et al., 2011). $Z_{true}$ is the real value of the depth of the points not used for training while the $Z_{predicted}$ is the predicted depth for these points, using the model trained on the rest of the points. The fitting score is calculated only in cases where a percentage of the dataset is used for training. Results in Table \ref{fig:t3} highlight the robustness of the proposed depth correction framework.
\section{CONCLUSIONS}\label{sec:CONCLUSIONS}
In the proposed approach, based on known depth observations from bathymetric LiDAR surveys, an SVR model was developed able to estimate with high accuracy the real depths of point clouds derived from conventional SfM-MVS procedures. Experimental results over two test sites along with the performed quantitative validation indicated the high potential of the developed approach and the wide field for machine and deep learning architectures in bathymetric applications. It is proved that the model can be trained on one area and used on another one, or indeed on many other, having different characteristics and achieving results of very high accuracy. The proposed approach can be used also in areas were LiDAR data of low density are available, in order to create a denser seabed representation. The methodology is independent from the UAV system used, also the camera and the flight height and there is no need for additional data i.e. camera orientations, camera intrinsic etc. for predicting the correct depth of a point cloud. This is a very important asset of the proposed method in relation to the other state of the art methods used for overcoming refraction errors in seabed mapping. The limitations of this method are mainly imposed by the SfM-MVS errors in areas having texture of low quality (e.g. sand and seagrass areas). Limitations are also imposed due to incompatibilities between the LiDAR point cloud and the image-based one. Among ohers, the different level of detail imposed additional errors in the point cloud comparison and compromise the absolute accuracy of the method. However, twelve out of thirteen different tests (Table \ref{fig:t3}) proved that the proposed method meets and exceeds the accuracy standards generally accepted for hydrography established by the International Hydrographic Organization (IHO), where in its simplest form, the vertical accuracy requirement for shallow water hydrography can be set as a total of $\pm$25cm (one sigma) from all sources, including tides \cite{Guenther2000}.
\section*{ACKNOWLEDGEMENTS}\label{ACKNOWLEDGEMENTS}
Authors would like to acknowledge the Dep. of Land and Surveys of Cyprus for providing the LiDAR reference data, and the Cyprus Dep. of Antiquities for permitting the flight over the Amathouda site and commissioning the flight over Ag. Napa. Also, authors would like to thank Dr. Ioannis Papadakis for the discussions on the physics of the refraction effect.
{
	\begin{spacing}{0.99}
		\normalsize
		\bibliography{ISPRSguidelines_authors} 
	\end{spacing}
}
\end{document}